\documentclass[sn-basic]{sn-jnl}

\usepackage{graphicx}
\usepackage{multirow}
\usepackage{amsmath,amssymb,amsfonts}
\usepackage{amsthm}
\usepackage{eucal}
\usepackage[title]{appendix}
\usepackage{xcolor}
\usepackage{textcomp}
\usepackage{manyfoot}
\usepackage{booktabs}
\usepackage{listings}
\usepackage[utf8]{inputenc}
\usepackage[T1]{fontenc}
\usepackage{geometry}
\usepackage{hyperref}
\usepackage{algorithm}
\usepackage{algpseudocode}
\usepackage{enumitem}

\theoremstyle{thmstyleone}%

\theoremstyle{thmstyletwo}%

\theoremstyle{thmstylethree}%
\definecolor{backcolour}{rgb}{0.95,0.95,0.92}
\definecolor{mygreen}{rgb}{0,0.6,0}
\definecolor{myblue}{rgb}{0.2,0.2,0.7}
\definecolor{mygray}{rgb}{0.5,0.5,0.5}

\begin{document}

\title{SplitWise Regression: Stepwise Modeling with Adaptive Dummy Encoding}

\author*[1]{\fnm{Marcell T.} \sur{Kurbucz}}\email{m.kurbucz@ucl.ac.uk}

\author[1]{\fnm{Nikolaos} \sur{Tzivanakis}}\email{n.tzivanakis@ucl.ac.uk}

\author[1]{\fnm{Nilufer Sari} \sur{Aslam}}\email{n.aslam.11@ucl.ac.uk}

\author[2]{\fnm{Adam M.} \sur{Sykulski}}\email{adam.sykulski@imperial.ac.uk}

\affil[1]{\orgdiv{Institute for Global Prosperity, The Bartlett}, \orgname{University College London}, \orgaddress{\street{149 Tottenham Court Road}, \city{London}, \postcode{W1T 7NE},  \country{United Kingdom}}}

\affil[2]{\orgdiv{Department of Mathematics, Faculty of Natural Sciences}, \orgname{Imperial College London}, \orgaddress{\street{180 Queen's Gate}, \city{London}, \postcode{SW7 2AZ}, \country{United Kingdom}}}

\abstract{
Capturing nonlinear relationships without sacrificing interpretability remains a persistent challenge in regression modeling. We introduce SplitWise, a novel framework that enhances stepwise regression. It adaptively transforms numeric predictors into threshold-based binary features using shallow decision trees, but only when such transformations improve model fit, as assessed by the Akaike Information Criterion (AIC) or Bayesian Information Criterion (BIC). This approach preserves the transparency of linear models while flexibly capturing nonlinear effects. Implemented as a user-friendly \textsf{R} package, \texttt{SplitWise} is evaluated on both synthetic and real-world datasets. The results show that it consistently produces more parsimonious and generalizable models than traditional stepwise and penalized regression techniques.
}


\maketitle

\flushbottom
\thispagestyle{empty}

\section{Introduction}

Statistical models that balance predictive performance with interpretability are increasingly valued across a range of disciplines. Traditional linear regression remains a popular choice due to its transparency: each coefficient directly quantifies the influence of a predictor on the outcome. However, its simplicity comes with limitations, particularly in its inability to capture nonlinear or threshold-based effects. By contrast, machine learning methods such as decision trees and neural networks can model complex relationships but often sacrifice interpretability in the process \citep{Breiman2001, Lipton2018}. In high-stakes domains such as healthcare, finance, and public policy, interpretability is not merely desirable---it is often essential for stakeholder trust and regulatory compliance \citep{Caruana2015, DoshiVelez2017}. This has driven interest in methods that aim to improve model flexibility while retaining interpretability \citep{Rudin2019}.

Stepwise regression \citep{Efroymson1960}, a longstanding technique for variable selection in linear models, systematically analyses various combinations of variables, adding or removing them based on their statistical importance \citep{Valiaho1969, ValiahoAndPekkonen1976}. This methodology typically employs three primary strategies: forward selection, which begins with an empty model and sequentially adds the most statistically significant variables; backward elimination, which starts with all candidate variables and progressively removes the least significant ones; and bidirectional elimination, which combines both approaches by allowing variables to be either added or removed at each step based on their statistical significance. Each iteration evaluates variables using F-statistics derived from t-tests on coefficient estimates, with the algorithm terminating when predefined statistical thresholds are satisfied \citep{MillerPanneerselvamLiu2022}. Despite criticisms regarding overfitting and statistical validity \citep{Harrell2001}, modern applications are focused on Akaike's Information Criterion (AIC) \citep{Akaike1974} and the Bayesian Information Criterion (BIC) \citep{Schwarz1978, Raftery1995}---which balance model fit and complexity, respectively. While current stepwise selection methods typically rely on modern applications, including ``a p-value threshold'' \citep{Fong2017}, there is a need to incorporate automatic detection of optimal thresholds in numeric variables to enhance model robustness and reduce the risk of overfitting.

Nevertheless, purely linear models may fail to capture meaningful nonlinearities or threshold effects present in real-world data. For instance, the relationship between age and health outcomes may change sharply after a certain age, or a risk factor may become relevant only beyond a critical level. Decision tree algorithms \citep{Breiman1984, Quinlan1993} address this by partitioning predictor space into regions via recursive splits, often yielding models that are piecewise constant or piecewise linear---such as in the M5 model tree algorithm \citep{Quinlan1992}. Recent variants have improved scalability and added regularization, including PILOT \citep{raymaekers2024fast}, which uses greedy splitting with ridge-regularized linear fits at the nodes, and MOTR-BART \citep{prado2021bayesian}, which integrates linear components into Bayesian additive regression trees. In applied domains such as cybersecurity, hybrid methods like logistic model trees have demonstrated the value of combining split-based structure with interpretable regression \citep{adeyemo2021ensemble}. While effective at capturing nonlinearities, these approaches typically result in complex model structures with multiple splits, nested conditions, or ensemble components, which diminish transparency and make it difficult to extract global insights---a fundamental limitation for applications where interpretability is as important as predictive performance.

Beyond tree-based models, rule-based hybrid approaches have emerged to balance interpretability with predictive power. RuleFit \citep{Friedman2008}, for example, constructs linear models from rule-based features extracted from decision trees. Such approaches can maintain a form of interpretability by expressing predictions as a linear combination of human-readable conditions \citep{Molnar2022}. However, they typically require training large ensembles and applying regularization to manage complexity, which can limit transparency and accessibility in practice. Furthermore, the resulting models often present users with numerous rules of varying importance, making it challenging to distill key insights about the underlying relationships between variables. A key distinction is that these methods do not specifically create or optimize binary threshold variables that directly enhance linear model performance while preserving the interpretability of the original feature space. This gap indicates a need for approaches that introduce nonlinearity in a more controlled and targeted manner---maintaining the coherence of a global linear regression while enabling the discovery of meaningful breakpoints in continuous variables.

In this paper, we propose SplitWise regression, a deterministic framework that integrates adaptive threshold-based transformations into a stepwise selection process. The method allows numeric variables to enter the model either as standard linear terms or as binary indicators based on data-driven cut-points. Each candidate transformation is evaluated using AIC or BIC to ensure that only interpretable, predictive modifications are retained. The resulting model remains a global linear equation, augmented by a small number of thresholded features---e.g., $I(18 < \text{age} < 34)$ or $I(\text{blood pressure} > 140)$---which preserve transparency while capturing relevant nonlinearity. Compared to full tree-based methods, this approach yields a compact and interpretable model that can be readily understood, audited, or used manually. To support adoption and reproducibility, we also release a user-friendly \textsf{R} package, \texttt{SplitWise}.

The remainder of this paper is structured as follows. Section~\ref{sec:methods} describes the SplitWise algorithm, software implementation, and evaluation datasets. Section~\ref{sec:results} presents experimental results and interpretation examples. Section~\ref{sec:conclusion} concludes and outlines directions for future research.

\section{Methodology, Software, and Data}
\label{sec:methods}

We begin by formally describing the SplitWise algorithm and its two transformation modes, followed by details on its software implementation and the datasets used for evaluation.

\subsection{SplitWise Regression Algorithm}

The SplitWise regression algorithm constructs interpretable regression models by automatically determining how each predictor should enter the model---either as a linear term or as one or more dummy-coded segments, derived via optimal split points. It uses information criteria (AIC or BIC) to balance model fit and complexity, ensuring that the selected transformations yield the best trade-off between interpretability and predictive power.

\begin{figure}[h]
    \centering
    \includegraphics[width=1\textwidth]{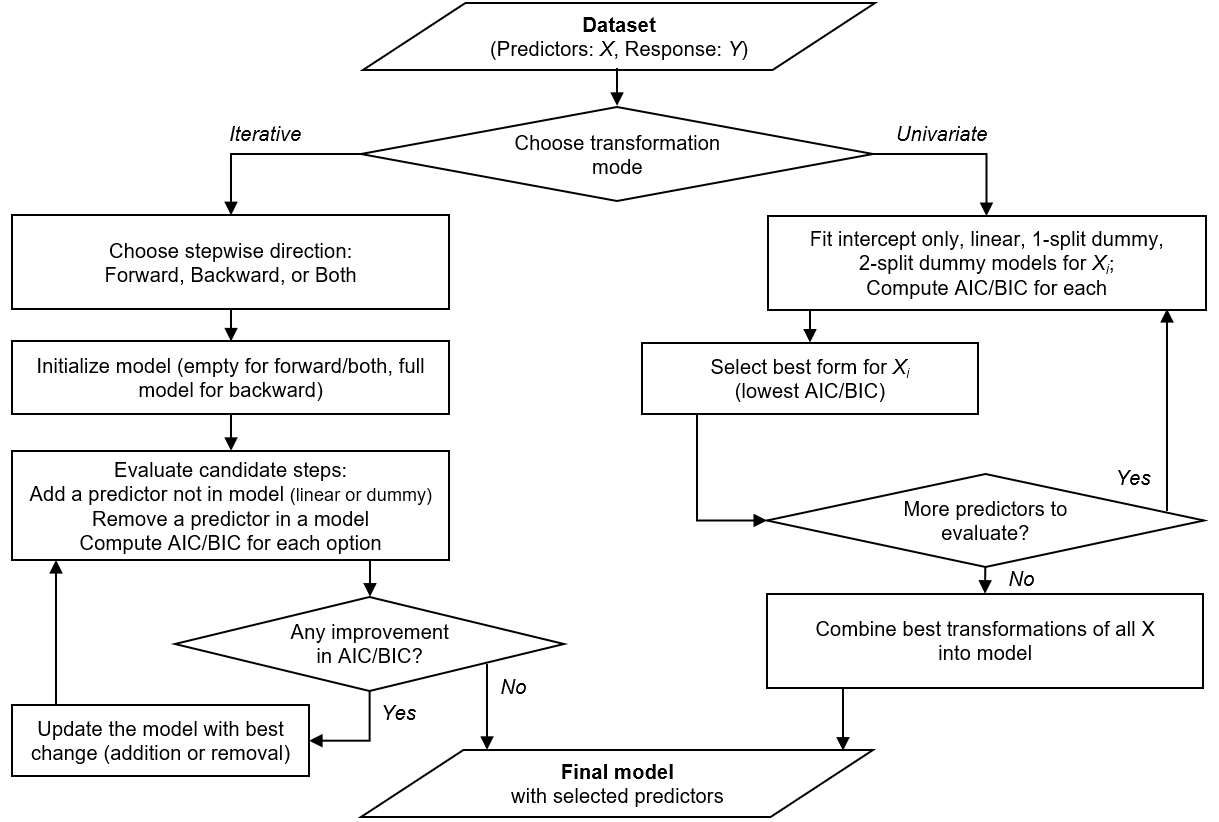}
    \caption{Workflow diagram of the SplitWise regression algorithm, illustrating the two transformation modes (iterative vs. univariate) and the sequence of steps for each.}
    \label{fig:splitwise_workflow}
\end{figure}

The algorithm operates in two modes---iterative and univariate---as outlined below and illustrated in Figure \ref{fig:splitwise_workflow}. The iterative mode is generally more precise, as it evaluates transformations in the context of other variables already included in the model, allowing it to account for variable interactions and joint effects. The univariate mode, by contrast, is more scalable to high-dimensional data, as it transforms each predictor independently. Although this step treats variables separately during transformation, a subsequent stepwise selection process is applied to assemble a well-performing final model. Each mode selects variables and transformations differently, but both produce a final regression model with the chosen predictors in their most appropriate form.

\noindent
\textit{A)} \underline{\textit{Iterative Mode:}}
\vspace{1em}

\noindent
In iterative mode, the algorithm builds the model via a stepwise procedure, adding or removing predictors one at a time based on AIC/BIC improvement. There are three strategies to control the search direction:

\begin{itemize}[left=0pt, label=\raisebox{0.15em}{\tiny$\bullet$}, itemsep=0.3em]
    \item \textbf{Forward selection:} Start with no predictors and iteratively add significant ones.
    \item \textbf{Backward elimination:} Start with a full model and iteratively remove non-informative predictors.
    \item \textbf{Bidirectional (stepwise):} Allow both addition and removal of variables at each step.
\end{itemize}

\noindent
Regardless of the strategy, the iterative process includes the following steps:

\begin{itemize}[left=0pt, label=\raisebox{0.15em}{\tiny$\bullet$}, itemsep=0.3em]
    \item \textbf{Initialization:} Start with a null model (forward/stepwise) or full model (backward), based on direction.
    
    \item \textbf{Generate candidates:}
    \begin{itemize}[label=--, left=1em, itemsep=0.0em]
        \item For addition: Evaluate each excluded predictor’s best form (linear or dummy-transformed via \texttt{rpart}) and compute AIC/BIC if added.
        \item For removal: Evaluate current predictors and compute AIC/BIC for the model if removed.
    \end{itemize}
    
    \item \textbf{Select best step:} Evaluate the best form for each excluded predictor (linear or dummy-transformed via \texttt{rpart}), where optimal split points are identified using binary recursive partitioning with a maximum depth of two, and pruning is applied using the complexity parameter. AIC/BIC are computed for each transformation, and the best model is selected based on the lowest criterion value.

    \item \textbf{Update model:} Implement the best action and repeat the loop.
    
    \item \textbf{Stop condition:} Terminate when no single action (add/remove/switch) improves AIC/BIC further. The current model is selected.
\end{itemize}

\noindent
\textit{B)} \underline{\textit{Univariate Mode:}}
\vspace{1em}

\noindent
In univariate mode, the algorithm evaluates each predictor independently to determine its most appropriate representation in the model. Similar to iterative mode, optimal split points are identified using the default settings of the \texttt{rpart} decision tree algorithm, which performs binary recursive partitioning with a maximum depth of two. A complexity parameter is also used to simplify the tree and prevent overfitting. The split points are selected to optimize model performance while minimizing overfitting, based on criteria such as reducing impurity (e.g., Gini index or variance).

\begin{itemize}[left=0pt, label=\raisebox{0.15em}{\tiny$\bullet$}, itemsep=0.3em]
    \item \textbf{Loop through predictors:} For each predictor $X_i$ in the dataset, fit multiple candidate models:
    \begin{itemize}[label=--, left=1em, itemsep=0.0em]
        \item Intercept-only model (null model, no effect of $X_i$).
        \item Linear model: $Y \sim X_i$.
        \item Single-split dummy model: Partition $X_i$ into two segments using an optimal split point from a shallow decision tree---e.g., \texttt{rpart} \citep{rpart}---and generate a 0/1 dummy variable.
        \item Double-split dummy model: Partition $X_i$ into three segments using two optimal split points, encoded via two dummy variables. This limit ensures parsimony while allowing detection of potential non-monotonic or threshold effects.
    \end{itemize}
    
    \item \textbf{Evaluate AIC/BIC:} Compute the information criterion (AIC or BIC) for each transformation. Lower values indicate better trade-offs between fit and complexity.
    
    \item \textbf{Select best transformation:} Choose the form of $X_i$ with the lowest AIC/BIC. If the null model is best, exclude $X_i$ from the final model.
    
    \item \textbf{Assemble final model:} Combine all selected predictors in their chosen forms into a final regression model. Excluded variables are omitted.
\end{itemize}

\noindent
\underline{\textit{Model Output:}}
\vspace{1em}

\noindent
Both modes yield a final regression model composed of a subset of predictors, each in its most appropriate representation. To support interpretability, the resulting model retains a linear structure while incorporating thresholded features when advantageous. In univariate mode, variables are evaluated independently and passed through stepwise selection; in iterative mode, transformation and selection are integrated at each step.

\subsection{Software Implementation}

The SplitWise algorithm has been implemented in a dedicated \textsf{R} package, \texttt{SplitWise}, available open-source via GitHub \citep{kurbucz2025git}. The package supports both transformation modes (\texttt{"iterative"} and \texttt{"univariate"}) through a single entry-point function: \texttt{splitwise()}. Key user options include the direction of stepwise selection (\texttt{"forward"}, \texttt{"backward"}, or \texttt{"both"}), the selection criterion (AIC or BIC), and control over which variables are eligible for transformation. To enhance interpretability and reproducibility, the returned model object (of class \texttt{"splitwise\_lm"}) includes metadata detailing the transformation decisions, cut-points for dummy variables, and the final design matrix. Custom \texttt{print()} and \texttt{summary()} methods are provided to visualize the model structure and transformations in a human-readable format, explicitly reporting which variables were dummy-encoded and how.

The package is designed to integrate seamlessly with existing \textsf{R} modeling workflows, relying primarily on base \textsf{R} functions such as \texttt{lm()} and \texttt{step()} for model fitting and selection. For threshold detection in numeric variables, it uses only a single external dependency: the \texttt{rpart} package \citep{rpart}. This minimal reliance on external libraries enhances robustness and ensures cross-platform compatibility---the \texttt{SplitWise} package is fully operating system-agnostic and functions identically across Windows, macOS, and Linux. Installation is handled via \texttt{devtools::install\_github("mtkurbucz/SplitWise")}, and the package includes a README file, examples, and complete function documentation. The software is released under the GPL-3 license. More information, including benchmark scripts, example datasets, and reproducible experiments, is available at: \url{https://github.com/mtkurbucz/SplitWise} (retrieved: April 27, 2025).

\vspace{1em}
\noindent
\underline{\textit{Minimal Code Example:}}
\vspace{1em}

\noindent
The following minimal example demonstrates how to apply the \texttt{SplitWise} package to the \textit{Mtcars} dataset \citep{henderson1974motortrend}, which contains data on various car models, including variables such as miles per gallon (\texttt{mpg}), horsepower, and weight. In this example, \texttt{mpg} is used as the target variable, and the method combines iterative transformations with backward stepwise selection. The resulting output includes model coefficients, residual diagnostics, dummy-encoded variable details, and information criteria (AIC and BIC).

\vspace{1em}

\begin{lstlisting}[language=R]
# Load the mtcars dataset
(*@\textcolor{myblue}{data}@*)(mtcars)

# Apply SplitWise with iterative transformations and backward stepwise selection
model <- (*@\textcolor{myblue}{splitwise}@*)(
  mpg ~ .,
  data = mtcars,
  transformation_mode = (*@\textcolor{gray}{"iterative"}@*),
  direction = (*@\textcolor{gray}{"backward"}@*)
)

# Display the summary of the fitted model
(*@\textcolor{myblue}{summary}@*)(model)
# Output:
# Call:
# stats::lm(formula = mpg ~ cyl + disp_dummy + hp + drat_dummy + am, data = df_final)

# Residuals:
# Min      1Q  Median      3Q     Max 
# -3.7656 -1.1218 -0.1794  1.2778  2.8054 

# Coefficients:
# (Intercept) 31.181550 (***)
# cyl         -0.819230 (*)
# disp_dummy  -6.542518 (***)
# hp          -0.029006 (**)
# drat_dummy   3.608055 (**)
# am           1.467368 

# Residual standard error: 1.71 on 26 degrees of freedom
# Multiple R-squared:  0.9325
# Adjusted R-squared:  0.9195
# F-statistic: 71.82 on 5 and 26 DF, p-value: 2.222e-14

# Dummy-Encoded Variables:
# - disp : 1 if x >= 101.550; else 0
# - drat : 1 if x >= 3.035; else 0
# - wt   : 1 if 1.885 < x < 3.013; else 0
# - qsec : 1 if 16.580 < x < 17.175; else 0
# - carb : 1 if 1.500 < x < 5.000; else 0

# Final AIC: 132.5
# Final BIC: 142.76
\end{lstlisting}

The final model includes an intercept term, two significant linear coefficients (both with p-values < 0.05), and two dummy-encoded variables, resulting in an easily interpretable representation of the SplitWise algorithm. The dummy-encoded variables are as follows: \texttt{disp\_dummy}, which indicates whether a car's engine displacement is greater than or equal to 101.55, and \texttt{drat\_dummy}, which captures whether the rear axle ratio exceeds 3.035. The positive coefficient for \texttt{drat\_dummy} suggests that cars with a higher rear axle ratio tend to exhibit better fuel efficiency (higher \texttt{mpg}), while the negative coefficient for \texttt{disp\_dummy} suggests that larger engine displacements are associated with lower fuel efficiency. These dummy variables enable the model to capture nonlinear effects in a flexible yet interpretable manner, maintaining both transparency and simplicity.

\subsection{Employed Datasets and Experimental Setup}

We evaluate SplitWise using both synthetic and real-world regression datasets. Each dataset is analyzed using a diverse suite of methods, including classical stepwise selection (in multiple directions), best subset selection via the \texttt{regsubsets()} function from the \texttt{leaps} package \citep{leaps}---which identifies the best model containing a given number of predictors---penalized regression techniques such as LASSO, Ridge, and Elastic Net \citep[see, e.g.,][]{altelbany2021evaluation}, the \texttt{dredge()} function from the \texttt{MuMIn} package \citep{MuMIn}---which performs automated model selection by generating all possible combinations of fixed-effect terms---and the proposed SplitWise framework. We deliberately focus on methods that produce directly interpretable linear models, and thus exclude complex machine learning techniques such as PILOT \citep{raymaekers2024fast}, ensemble methods, or deep neural networks, which prioritize predictive power over interpretability. Performance is assessed across several criteria: root mean squared error (RMSE), mean absolute error (MAE), adjusted $R^2$, AIC, BIC, number of predictors selected, computation time, and the stability of variable selection---measured by the proportion of times the same set of predictors is selected across different resampling runs.

This comprehensive benchmarking framework ensures consistent comparisons among interpretable methods and illustrates how each technique scales with increasing dimensionality and complexity. The remainder of this section describes the datasets used.

\vspace{1em}
\noindent
\underline{\textit{Synthetic Datasets:}}
\vspace{1em}

\noindent
Synthetic datasets were generated using the \texttt{simstudy} package \citep{simstudy} in \textsf{R}. Each dataset consisted of 1,000 observations with varying numbers of predictors (15, 30, and 60 predictors). To reflect realistic challenges in regression modeling, the datasets included multicollinearity (by correlating selected predictors), added noise, and sparsity (with only a small subset of predictors contributing to the response). For each predictor size, 100 independent datasets were generated with different random seeds to allow for robust performance evaluation and variability estimation. The true underlying response was constructed as a sparse linear combination of a few predictors, while the remaining variables served as noise. This design facilitates a thorough evaluation of each method’s ability to recover meaningful signals under high-dimensional, noisy conditions; however, due to its high computational cost, the dredge method was not applied in the synthetic experiments.

\vspace{1em}
\noindent
\underline{\textit{Real-World Datasets:}}
\vspace{1em}

\noindent
To evaluate generalizability, we apply all methods to five publicly available datasets spanning health, housing, automotive engineering, and food quality. These datasets differ in size, complexity, and domain context, providing a broad testbed for assessing model robustness. A summary is provided in Table~\ref{tab:datasets}.

\begin{table}[ht]
\centering
\resizebox{0.85\textwidth}{!}{%
\begin{tabular}{lcccc}
\toprule
\textbf{Dataset} & \textbf{Sample Size} & \textbf{Predictors} & \textbf{Target} & \textbf{Source} \\
\midrule
Bodyfat & 252 & 14 & Body Fat (\%) & a \\
Boston Housing & 506 & 13 & Home Value (\$1000s) & b \\
Mtcars & 32 & 10 & Miles per gallon (mpg) & c \\
Wine Quality (Red) & 1,599 & 11 & Quality score (0--10) & d \\
Wine Quality (White) & 4,898 & 11 & Quality score (0--10) & d \\
\bottomrule
\end{tabular}%
}
\caption{Summary of real-world datasets used for evaluating regression methods. \textbf{Sources:}
a: \cite{johnson1996bodyfat};
b: \cite{harrison1978hedonic};
c: \cite{henderson1974motortrend};
d: \cite{cortez2009wine}.
}
\label{tab:datasets}
\end{table}

Note that, in contrast to the synthetic experiments where performance statistics were averaged over multiple independently generated datasets, the real-world datasets consist of a single fixed sample. Consequently, standard deviations are not reported for real-world results, ensuring a direct and unbiased comparison across methods.

\section{Results and Discussion}
\label{sec:results}

This section presents the empirical evaluation of regression methods using synthetic and real-world datasets. We first analyze model performance in controlled, simulated settings, then assess generalizability on real-world data.

\subsection{Synthetic Datasets}

Table~\ref{tab:synthetic-results} summarizes the comparative performance of all methods across synthetic datasets with varying numbers of predictors (15, 30, and 60).

\begin{table}[htp]
\centering
\resizebox{0.95\textwidth}{!}{%
\begin{tabular}{llcccccccc}
\toprule
\textbf{Predictors} & \textbf{Settings} & \textbf{AIC} & \textbf{BIC} & \textbf{Adj. $R^2$} & \textbf{RMSE} & \textbf{MAE} & \textbf{Stability} & \textbf{Vars} & \textbf{Time (s)} \\
\midrule
\multicolumn{10}{l}{\textbf{15 Predictors}} \\
SplitWise & iter.; backw. & \textbf{4433 (47)} & \textbf{4474 (47)} & \textbf{0.580 (0.02)} & \textbf{2.190 (0.05)} & \textbf{1.746 (0.04)} & \textbf{1.00} & 5.34 (1.51) & 1.314 \\
SplitWise & iter.; forw. & 4435 (47) & 4474 (47) & 0.578 (0.02) & 2.195 (0.05) & 1.751 (0.04) & \textbf{1.00} & 4.95 (1.46) & 0.415 \\
SplitWise & iter.; both & 4435 (47) & 4473 (47) & 0.578 (0.02) & 2.196 (0.05) & 1.751 (0.04) & \textbf{1.00} & 4.77 (1.42) & 0.589 \\
SplitWise & univ.; forw. & 4460 (46) & 4549 (46) & 0.568 (0.02) & 2.210 (0.05) & 1.762 (0.05) & \textbf{1.00} & 16.41 (1.04) & 0.054 \\
SplitWise & univ.; backw. & 4444 (46) & 4480 (46) & 0.570 (0.02) & 2.216 (0.05) & 1.767 (0.05) & \textbf{1.00} & 4.45 (1.34) & 0.418 \\
SplitWise & univ.; both & 4444 (46) & 4480 (46) & 0.570 (0.02) & 2.216 (0.05) & 1.767 (0.05) & \textbf{1.00} & 4.45 (1.34) & 0.221 \\
Stepwise & forw. & 4461 (46) & 4549 (46) & 0.567 (0.02) & 2.212 (0.05) & 1.763 (0.05) & \textbf{1.00} & 15.00 (0) & 0.005 \\
Stepwise & backw. & 4445 (46) & 4481 (46) & 0.570 (0.02) & 2.217 (0.05) & 1.768 (0.05) & 0.06 & 4.30 (1.31) & 0.217 \\
Stepwise & both & 4445 (46) & 4481 (46) & 0.570 (0.02) & 2.217 (0.05) & 1.768 (0.05) & 0.06 & 4.30 (1.31) & 0.153 \\
Best Subset & size: 3--4 & 4448 (46) & 4473 (46) & 0.567 (0.02) & 2.226 (0.05) & 1.775 (0.05) & 0.59 & 2.11 (0.31) & 0.014 \\
Ridge & -- & -- & -- & -- & 2.219 (0.05) & 1.769 (0.05) & -- & 15 (0) & 0.079 \\
LASSO & -- & -- & -- & -- & 2.221 (0.05) & 1.771 (0.05) & -- & 6.38 (2.53) & 0.091 \\
Elastic Net & -- & -- & -- & -- & 2.219 (0.05) & 1.769 (0.05) & -- & 8.56 (2.61) & 0.053 \\
\midrule
\multicolumn{10}{l}{\textbf{30 Predictors}} \\
SplitWise & iter.; backw. & \textbf{4428 (44)} & \textbf{4483 (43)} & \textbf{0.600 (0.02)} & \textbf{2.124 (0.06)} & \textbf{1.694 (0.05)} & \textbf{1.00} & 8.19 (2.24) & 4.497 \\
SplitWise & iter.; forw. & 4432 (44) & 4485 (43) & 0.596 (0.02) & 2.135 (0.06) & 1.703 (0.05) & \textbf{1.00} & 7.83 (2.46) & 1.327 \\
SplitWise & iter.; both & 4431 (43) & 4482 (42) & 0.596 (0.02) & 2.137 (0.06) & 1.706 (0.05) & \textbf{1.00} & 7.26 (2.17) & 1.784 \\
SplitWise & univ.; forw. & 4481 (43) & 4646 (43) & 0.567 (0.02) & 2.200 (0.05) & 1.754 (0.04) & \textbf{1.00} & 32.54 (1.49) & 0.092 \\
SplitWise & univ.; backw. & 4446 (43) & 4494 (43) & 0.572 (0.02) & 2.213 (0.05) & 1.765 (0.04) & \textbf{1.00} & 6.89 (2.15) & 0.596 \\
SplitWise & univ.; both & 4446 (43) & 4494 (43) & 0.572 (0.02) & 2.213 (0.05) & 1.765 (0.04) & \textbf{1.00} & 6.89 (2.15) & 0.818 \\
Stepwise & forw. & 4485 (42) & 4647 (42) & 0.565 (0.02) & 2.205 (0.05) & 1.759 (0.04) & \textbf{1.00} & 30.00 (0) & 0.007 \\
Stepwise & backw. & 4450 (42) & 4496 (43) & 0.570 (0.02) & 2.218 (0.05) & 1.770 (0.04) & 0.01 & 6.40 (1.97) & 0.441 \\
Stepwise & both & 4450 (42) & 4496 (43) & 0.570 (0.02) & 2.218 (0.05) & 1.770 (0.04) & 0.01 & 6.40 (1.97) & 0.684 \\
Best Subset & size: 3--5 & 4455 (42) & 4481 (42) & 0.566 (0.02) & 2.234 (0.05) & 1.782 (0.04) & 0.54 & 2.19 (0.44) & 0.069 \\
Ridge & -- & -- & -- & -- & 2.212 (0.05) & 1.764 (0.04) & -- & 30 (0) & 0.061 \\
LASSO & -- & -- & -- & -- & 2.229 (0.05) & 1.778 (0.04) & -- & 8.41 (4.56) & 0.053 \\
Elastic Net & -- & -- & -- & -- & 2.227 (0.05) & 1.776 (0.04) & -- & 11.11 (4.30) & 0.049 \\
\midrule
\multicolumn{10}{l}{\textbf{60 Predictors}} \\
SplitWise & iter.; backw. & \textbf{4381 (56)} & \textbf{4473 (52)} & \textbf{0.600 (0.02)} & \textbf{2.124 (0.06)} & \textbf{1.694 (0.05)} & \textbf{1.00} & 15.65 (3.67) & 33.033 \\
SplitWise & iter.; forw. & 4390 (56) & 4479 (52) & 0.596 (0.02) & 2.135 (0.06) & 1.703 (0.05) & \textbf{1.00} & 15.03 (3.52) & 9.267 \\
SplitWise & iter.; both & 4389 (56) & 4468 (52) & 0.596 (0.02) & 2.137 (0.06) & 1.706 (0.05) & \textbf{1.00} & 13.18 (3.00) & 11.486 \\
SplitWise & univ.; forw. & 4499 (53) & 4813 (52) & 0.570 (0.02) & 2.153 (0.06) & 1.719 (0.05) & \textbf{1.00} & 65.15 (1.87) & 0.176 \\
SplitWise & univ.; backw. & 4427 (52) & 4502 (50) & 0.580 (0.02) & 2.181 (0.06) & 1.741 (0.05) & \textbf{1.00} & 12.24 (3.21) & 5.012 \\
SplitWise & univ.; both & 4427 (52) & 4502 (50) & 0.580 (0.02) & 2.180 (0.06) & 1.741 (0.05) & \textbf{1.00} & 12.29 (3.27) & 7.029 \\
Stepwise & forw. & 4505 (51) & 4814 (51) & 0.566 (0.02) & 2.162 (0.06) & 1.726 (0.05) & \textbf{1.00} & 60.00 (0) & 0.013 \\
Stepwise & backw. & 4433 (50) & 4504 (50) & 0.577 (0.02) & 2.189 (0.06) & 1.748 (0.05) & 0.01 & 11.42 (3.35) & 4.030 \\
Stepwise & both & 4433 (50) & 4504 (50) & 0.577 (0.02) & 2.189 (0.06) & 1.748 (0.05) & 0.01 & 11.45 (3.38) & 5.647 \\
Best Subset & size: 3--6 & 4444 (49) & 4471 (49) & 0.569 (0.02) & 2.221 (0.06) & 1.773 (0.05) & 0.54 & 2.54 (0.70) & 156.721 \\
Ridge & -- & -- & -- & -- & 2.169 (0.06) & 1.732 (0.05) & -- & 60 (0) & 0.102 \\
LASSO & -- & -- & -- & -- & 2.218 (0.06) & 1.770 (0.05) & -- & 9.87 (5.83) & 0.067 \\
Elastic Net & -- & -- & -- & -- & 2.215 (0.06) & 1.768 (0.05) & -- & 14.19 (6.48) & 0.062 \\
\bottomrule
\end{tabular}
}
\caption{Performance summary of regression methods across synthetic datasets with 15, 30, and 60 predictors. All values are reported as means with standard deviations shown in parentheses. Best results for AIC, BIC, RMSE, adj. $R^2$, and stability are bolded. Stability measures the proportion of times the same set of predictors was selected across different resampling runs, with a value of 1.00 indicating perfect reproducibility. Stability is reported before the average number of selected variables (Vars); note that the mean number of variables may differ slightly across repetitions even when stability is 1.00, due to coefficient shrinkage effects or thresholding of near-zero coefficients.}
\label{tab:synthetic-results}
\end{table}

As presented in Table~\ref{tab:synthetic-results}, the results across synthetic datasets emphasize the consistent advantages of the proposed SplitWise approach---particularly in its iterative configurations. SplitWise achieved the lowest or tied-lowest RMSE and the highest adjusted $R^2$ in every setting, demonstrating its ability to maintain high predictive accuracy while avoiding overfitting. In both the 15- and 30-predictor scenarios, it also achieved the lowest AIC and BIC values, further supporting its effectiveness in balancing model fit and complexity. Even in the high-dimensional 60-predictor case, SplitWise sustained top-tier performance, with its iterative variants clearly outperforming other methods on all key metrics except execution time. Unlike the exhaustive search approach of best subset selection, SplitWise maintains feasible runtimes even in moderately high-dimensional settings.

A key strength of SplitWise lies in its ability to identify very sparse yet expressive models. For instance, in the 15-predictor setting, its iterative configurations achieved optimal accuracy with as few as four to five variables on average---compared to stepwise approaches and penalized models, which often required substantially more predictors for similar or worse performance. The method also demonstrated robust consistency across different initialization strategies (forward, backward, and both), underscoring its stability under various search directions. However, it is notable that the univariate forward variant of SplitWise tended to select substantially more predictors, particularly in higher-dimensional settings, similar to the behavior observed in traditional stepwise forward selection. This highlights a trade-off shared by both approaches: while forward search strategies maintain strong predictive accuracy, they often do so at the expense of model sparsity, potentially limiting interpretability compared to their iterative or backward-based counterparts.

Traditional stepwise regression remained competitive in lower dimensions but generally selected larger models and did not surpass SplitWise in any key metric. Its limitations became more apparent as dimensionality increased, with diminishing gains in accuracy and less favorable model compactness. Penalized regression methods like LASSO, Ridge, and Elastic Net were computationally efficient but consistently lagged behind in RMSE and adjusted $R^2$. Moreover, as continuous shrinkage methods, penalized regressions do not permit meaningful evaluation of stability based on variable selection consistency, and they tended to retain larger models.

\subsection{Real-World Datasets}

Table~\ref{tab:realworld-results} summarizes the results obtained on the real-world datasets.

\begin{table}[htp]
\centering
\resizebox{0.8\textwidth}{!}{%
\begin{tabular}{llccccccc}
\toprule
\textbf{Dataset} & \textbf{Settings} & \textbf{AIC} & \textbf{BIC} & \textbf{Adj. $R^2$} & \textbf{RMSE} & \textbf{MAE} & \textbf{Vars} & \textbf{Time (s)} \\
\midrule
\multicolumn{9}{l}{\textbf{Bodyfat}} \\
SplitWise & iter.; backw. & 377.48 & 391.06 & 0.911 & 3.174 & 2.229 & 3 & 0.560 \\
SplitWise & iter.; forw. & 382.90 & 407.79 & 0.910 & \textbf{3.073} & 2.065 & 8 & 0.507 \\
SplitWise & iter.; both & 377.92 & 393.75 & 0.912 & 3.139 & 2.198 & 4 & 0.757 \\
SplitWise & univ.; backw. & \textbf{373.58} & \textbf{387.16} & \textbf{0.916} & 3.088 & 2.275 & 3 & 0.068 \\
SplitWise & univ.; forw. & 383.25 & 410.41 & 0.910 & 3.037 & 2.186 & 9 & 0.042 \\
SplitWise & univ.; both & \textbf{373.58} & \textbf{387.16} & \textbf{0.916} & 3.088 & 2.275 & 3 & 0.088 \\
Stepwise & backw. & \textbf{373.58} & \textbf{387.16} & \textbf{0.916} & 3.088 & 2.275 & 3 & 0.064 \\
Stepwise & forw. & 381.41 & 406.30 & 0.912 & 3.041 & 2.184 & 8 & 0.004 \\
Stepwise & both & \textbf{373.58} & \textbf{387.16} & \textbf{0.916} & 3.088 & 2.275 & 3 & 0.031 \\
Best Subset & size: 4 & \textbf{373.58} & \textbf{387.16} & \textbf{0.916} & 3.088 & 2.275 & 3 & 0.004 \\
Dredge & best AICc model & \textbf{373.58} & \textbf{387.16} & \textbf{0.916} & 3.088 & 2.275 & 3 & 0.858 \\
LASSO & $\lambda$: 0.2627 & -- & -- & -- & 3.093 & 2.252 & 6 & 0.038 \\
Ridge & $\lambda$: 1.7287 & -- & -- & -- & 3.143 & 2.344 & 8 & 0.045 \\
Elastic Net & $\alpha$: 0.5, $\lambda$: 0.4788 & -- & -- & -- & 3.098 & 2.277 & 6 & 0.040 \\
\midrule
\multicolumn{9}{l}{\textbf{Boston Housing}} \\
SplitWise & iter.; backw. & 2969.93 & 3024.87 & 0.762 & 4.437 & 3.092 & 10 & 1.215 \\
SplitWise & iter.; forw. & 2896.82 & 2960.22 & \textbf{0.794} & \textbf{4.112} & \textbf{2.907} & 12 & 1.203 \\
SplitWise & iter.; both & \textbf{2894.55} & \textbf{2949.49} & \textbf{0.795} & 4.119 & 2.922 & 10 & 2.102 \\
SplitWise & univ.; backw. & 3016.30 & 3075.47 & 0.739 & 4.636 & 3.263 & 11 & 0.132 \\
SplitWise & univ.; forw. & 3018.82 & 3086.45 & 0.739 & 4.630 & 3.268 & 15 & 0.066 \\
SplitWise & univ.; both & 3016.30 & 3075.47 & 0.739 & 4.636 & 3.263 & 11 & 0.134 \\
Stepwise & backw. & 3023.73 & 3078.67 & 0.735 & 4.680 & 3.272 & 10 & 0.059 \\
Stepwise & forw. & 3027.61 & 3091.01 & 0.734 & 4.679 & 3.271 & 12 & 0.006 \\
Stepwise & both & 3023.73 & 3078.67 & 0.735 & 4.680 & 3.272 & 10 & 0.028 \\
Best Subset & size: 10 & 3032.00 & 3082.72 & 0.730 & 4.727 & 3.290 & 9 & 0.006 \\
Dredge & best AICc model & 3023.73 & 3078.67 & 0.735 & 4.680 & 3.272 & 10 & 17.290 \\
LASSO & $\lambda$: 0.0255 & -- & -- & -- & 4.683 & 3.258 & 10 & 0.052 \\
Ridge & $\lambda$: 0.6778 & -- & -- & -- & 4.734 & 3.235 & 12 & 0.070 \\
Elastic Net & $\alpha$: 0.5, $\lambda$: 0.0424 & -- & -- & -- & 4.683 & 3.257 & 10 & 0.051 \\
\midrule
\multicolumn{9}{l}{\textbf{Mtcars}} \\
SplitWise & iter.; backw. & \textbf{132.50} & \textbf{142.76} & \textbf{0.920} & \textbf{1.541} & \textbf{1.290} & 4 & 0.481 \\
SplitWise & iter.; forw. & 153.74 & 166.93 & 0.851 & 2.018 & 1.648 & 6 & 0.453 \\
SplitWise & iter.; both & 150.69 & 158.02 & 0.850 & 2.180 & 1.604 & 2 & 0.719 \\
SplitWise & univ.; backw. & 154.12 & 161.45 & 0.834 & 2.300 & 1.932 & 2 & 0.083 \\
SplitWise & univ.; forw. & 164.68 & 183.73 & 0.803 & 2.113 & 1.678 & 11 & 0.041 \\
SplitWise & univ.; both & 154.12 & 161.45 & 0.834 & 2.300 & 1.932 & 2 & 0.097 \\
Stepwise & backw. & 154.12 & 161.45 & 0.834 & 2.300 & 1.932 & 2 & 0.079 \\
Stepwise & forw. & 163.71 & 181.30 & 0.807 & 2.147 & 1.723 & 9 & 0.003 \\
Stepwise & both & 154.12 & 161.45 & 0.834 & 2.300 & 1.932 & 2 & 0.044 \\
Best Subset & size: 3 & 154.12 & 161.45 & 0.834 & 2.300 & 1.932 & 2 & 0.005 \\
Dredge & best AICc model & 154.12 & 161.45 & 0.834 & 2.300 & 1.932 & 2 & 1.684 \\
LASSO & $\lambda$: 0.8007 & -- & -- & -- & 2.509 & 1.966 & 2 & 0.054 \\
Ridge & $\lambda$: 2.7468 & -- & -- & -- & 2.320 & 1.848 & 9 & 0.067 \\
Elastic Net & $\alpha$: 0.5, $\lambda$: 0.7608 & -- & -- & -- & 2.320 & 1.847 & 7 & 0.048 \\
\midrule
\multicolumn{9}{l}{\textbf{Wine Quality (Red)}} \\
SplitWise & iter.; backw. & 3103.67 & 3152.06 & 0.379 & 0.635 & 0.496 & 6 & 0.783 \\
SplitWise & iter.; forw. & \textbf{3094.48} & \textbf{3148.25} & \textbf{0.382} & \textbf{0.633} & \textbf{0.493} & 7 & 0.581 \\
SplitWise & iter.; both & \textbf{3094.48} & \textbf{3148.25} & \textbf{0.382} & \textbf{0.633} & \textbf{0.493} & 7 & 0.905 \\
SplitWise & univ.; backw. & 3158.98 & 3207.37 & 0.357 & 0.646 & 0.501 & 6 & 0.183 \\
SplitWise & univ.; forw. & 3166.24 & 3241.52 & 0.356 & 0.646 & 0.500 & 16 & 0.056 \\
SplitWise & univ.; both & 3158.98 & 3207.37 & 0.357 & 0.646 & 0.501 & 6 & 0.235 \\
Stepwise & backw. & 3158.98 & 3207.37 & 0.357 & 0.646 & 0.501 & 6 & 0.092 \\
Stepwise & forw. & 3164.28 & 3234.18 & 0.356 & 0.646 & 0.500 & 10 & 0.004 \\
Stepwise & both & 3158.98 & 3207.37 & 0.357 & 0.646 & 0.501 & 6 & 0.055 \\
Best Subset & size: 6 & 3162.70 & 3205.72 & 0.355 & 0.647 & 0.503 & 5 & 0.006 \\
Dredge & best AICc model & 3158.98 & 3207.37 & 0.357 & 0.646 & 0.501 & 6 & 5.111 \\
LASSO & $\lambda$: 6e\!-\!04 & -- & -- & -- & 0.646 & 0.501 & 10 & 0.067 \\
Ridge & $\lambda$: 0.0384 & -- & -- & -- & 0.646 & 0.502 & 10 & 0.066 \\
Elastic Net & $\alpha$: 0.5, $\lambda$: 0.0186 & -- & -- & -- & 0.647 & 0.504 & 8 & 0.066 \\
\midrule
\multicolumn{9}{l}{\textbf{Wine Quality (White)}} \\
SplitWise & iter.; backw. & 10998.05 & 11082.51 & 0.297 & 0.742 & 0.577 & 10 & 0.926 \\
SplitWise & iter.; forw. & 11225.48 & 11303.44 & 0.263 & 0.759 & 0.594 & 9 & 1.514 \\
SplitWise & iter.; both & 11222.76 & 11281.23 & 0.263 & 0.759 & 0.594 & 6 & 2.419 \\
SplitWise & univ.; backw. & \textbf{10972.83} & \textbf{11070.28} & \textbf{0.301} & \textbf{0.739} & \textbf{0.574} & 12 & 0.478 \\
SplitWise & univ.; forw. & 10974.10 & 11078.05 & \textbf{0.301} & \textbf{0.739} & \textbf{0.574} & 18 & 0.125 \\
SplitWise & univ.; both & \textbf{10972.83} & \textbf{11070.28} & \textbf{0.301} & \textbf{0.739} & \textbf{0.574} & 12 & 0.664 \\
Stepwise & backw. & 11108.29 & 11173.25 & 0.281 & 0.750 & 0.584 & 7 & 0.119 \\
Stepwise & forw. & 11113.48 & 11197.94 & 0.280 & 0.750 & 0.584 & 10 & 0.005 \\
Stepwise & both & 11108.29 & 11173.25 & 0.281 & 0.750 & 0.584 & 7 & 0.114 \\
Best Subset & size: 8 & 11108.29 & 11173.25 & 0.281 & 0.750 & 0.584 & 7 & 0.012 \\
Dredge & best AICc model & 11108.29 & 11173.25 & 0.281 & 0.750 & 0.584 & 7 & 10.065 \\
LASSO & $\lambda$: 0.0064 & -- & -- & -- & 0.752 & 0.586 & 8 & 0.098 \\
Ridge & $\lambda$: 0.0386 & -- & -- & -- & 0.752 & 0.585 & 10 & 0.127 \\
Elastic Net & $\alpha$: 0.5, $\lambda$: 0.0042 & -- & -- & -- & 0.751 & 0.584 & 9 & 0.362 \\
\bottomrule
\end{tabular}
}
\caption{Performance of regression methods across real-world datasets. The best-performing models in each dataset are shown in bold.}
\label{tab:realworld-results}
\end{table}

SplitWise, especially in its iterative configuration, emerged as one of the most effective model selection strategies across all evaluated datasets. In complex datasets such as \textit{Boston Housing}, \textit{Wine Quality (Red)}, and \textit{Wine Quality (White)}, SplitWise consistently achieved the lowest RMSE. This suggests strong predictive performance even under challenging conditions. Notably, SplitWise reached this level of accuracy while selecting relatively small subsets of predictors---often fewer than or comparable to traditional methods---highlighting its ability to produce sparse yet expressive models. For example, in the \textit{Wine Quality (Red)} dataset, SplitWise (iterative) not only outperformed other methods in RMSE but did so with just 7 variables, avoiding the over-selection tendencies often observed in penalized models. Similarly, in the high-dimensional \textit{Wine Quality (White)} dataset, SplitWise matched the best-performing configurations while selecting fewer variables and avoiding the instability seen in methods like best subset selection or dredge (\texttt{MuMIn}). These patterns suggest that SplitWise effectively balances the competing goals of accuracy and interpretability, an important feature for real-world modeling scenarios where both are critical.

While traditional stepwise regression and best subset methods remained competitive---especially in lower-dimensional datasets like \textit{Bodyfat} and \textit{Mtcars}---their advantages were often limited to very specific cases and did not consistently generalize across datasets. They also showed signs of inefficiency or instability when applied to more complex datasets, where SplitWise continued to perform reliably. Penalized regression methods such as LASSO, Ridge, and Elastic Net offered efficient predictions but typically did not achieve the same level of RMSE performance and often retained a larger number of predictors. Moreover, methods like dredge (\texttt{MuMIn}), while occasionally matching traditional techniques in performance, incurred high computational costs, especially on larger datasets.

\section{Conclusion and Future Work}
\label{sec:conclusion}

We presented SplitWise, a novel and interpretable regression framework that enhances classical stepwise modeling through adaptive threshold-based dummy encoding. This method consistently outperformed traditional stepwise and modern penalized regression approaches across both synthetic and real-world datasets, particularly in terms of prediction accuracy and model parsimony. By blending model selection rigor with interpretability and minimal dependencies, its \textsf{R} implementation, \texttt{SplitWise}, offers a practical tool for analysts in fields such as healthcare, economics, and social sciences, where both high performance and interpretability are equally critical.

In controlled experiments, SplitWise reliably recovered sparse, high-accuracy models---even in high-dimensional settings---outperforming alternatives like LASSO and Ridge regression. In real-world applications, it maintained this advantage, particularly in datasets with complex or nonlinear interactions that standard linear models often struggle to capture. Its use of data-driven, threshold-based encoding within a stepwise framework enables it to flexibly model nonlinearity while preserving a globally interpretable linear form.

Looking forward, several promising extensions remain, which aim to expand the scope of SplitWise while maintaining its core strengths. A key priority is developing a \textsf{Python} implementation to support wider adoption and integration with modern data science workflows. Additionally, future work will explore hybrid strategies that combine SplitWise’s threshold-based dummy variable construction with more advanced, interpretable variable selection techniques---such as regularized stepwise search---to further enhance performance and sparsity. We also aim to develop accelerated variants of SplitWise to maintain its practical runtime performance even in ultra-high-dimensional settings \citep[see, e.g.,][]{hanczar2023feature}. Finally, extensions to generalized linear models (e.g., logistic regression) and time series forecasting are planned. Throughout, a central goal will remain: expanding the scope of SplitWise while preserving its strong interpretability and usability.

\bibliography{sn-bibliography}

%\section*{Acknowledgements}

%Acknowledgements should be brief, and should not include thanks to anonymous referees and editors, or effusive comments. Grant or contribution numbers may be acknowledged.

%\section*{Author contributions statement}

%Must include all authors, identified by initials, for example: A.A. conceived the experiment(s),  A.A. and B.A. conducted the experiment(s), C.A. and D.A. analysed the results.  All authors reviewed the manuscript.

\end{document}